\documentclass{article}

\PassOptionsToPackage{numbers, compress}{natbib}
\usepackage[]{neurips_2019}

\usepackage[utf8]{inputenc} 
\usepackage[T1]{fontenc}    
\usepackage{hyperref}       
\usepackage{url}            
\usepackage{booktabs}       
\usepackage{amsfonts}       
\usepackage{nicefrac}       
\usepackage{microtype}      
\usepackage{graphicx} 
\usepackage{subcaption}

\title{Learning from a Teacher using Unlabeled Data}

\begin{document}

\maketitle

\begin{abstract}
  Knowledge distillation is a widely used technique for model compression. We posit that the teacher model used in a distillation setup, captures relationships between classes, that extend beyond the original dataset. We empirically show that a teacher model can transfer this knowledge to a student model even on an {\it out-of-distribution} dataset. Using this approach, we show promising results on MNIST, CIFAR-10, and Caltech-256 datasets using unlabeled image data from different sources. Our results are encouraging and help shed further light from the perspective of understanding knowledge distillation and utilizing unlabeled data to improve model quality. 
\end{abstract}

\section{Introduction}
\begin{figure}[h]
  \centering
  \includegraphics[width=0.75\linewidth, height=5.0cm]{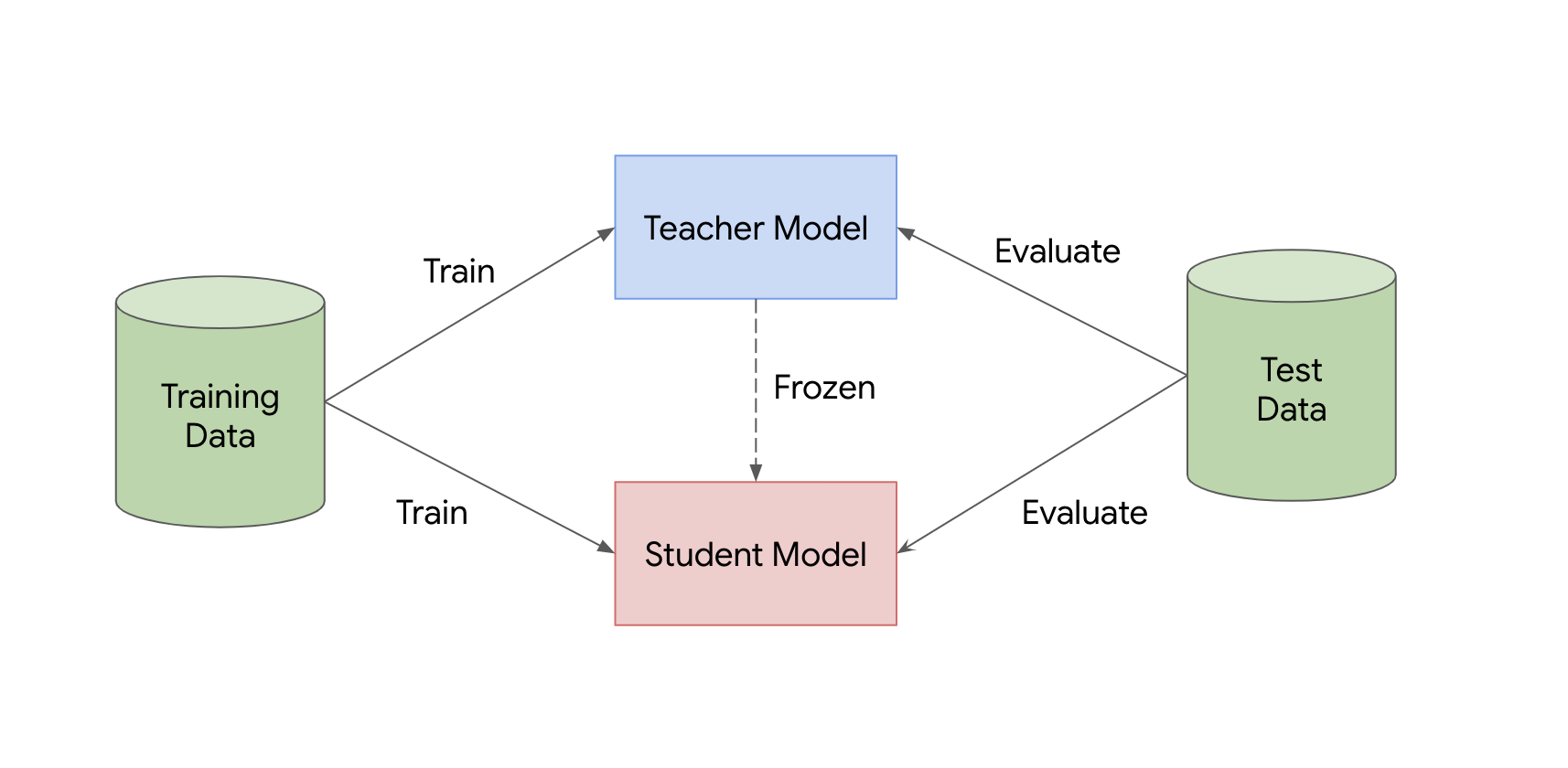}
  \caption{Vanilla Distillation: One or more teacher models are trained with the training data. The student model then learns to mimic the predictions of the \textit{frozen} teacher model by treating them as `soft labels'.}
  \label{vanilla-distillation}
\end{figure}

Knowledge distillation was introduced by Hinton et al. \citep{Hinton14} primarily for model compression by introducing a method to replicate the performance of model ensembles with a single model. One or more teacher models can be used to distill information into a single student model, generally of lower capacity in terms of model parameters. The teacher models help by providing soft-labels to the student, which allow it to learn a function to better discriminate between the different classes (Figure~\ref{vanilla-distillation}).

Our core hypothesis is that a teacher model can help the student model discriminate between the classes using not just the original training data, but also through a large corpus of unlabeled data available from a different distribution.


\begin{figure}[h]
  \centering
  \includegraphics[width=0.6\linewidth, height=4.0cm]{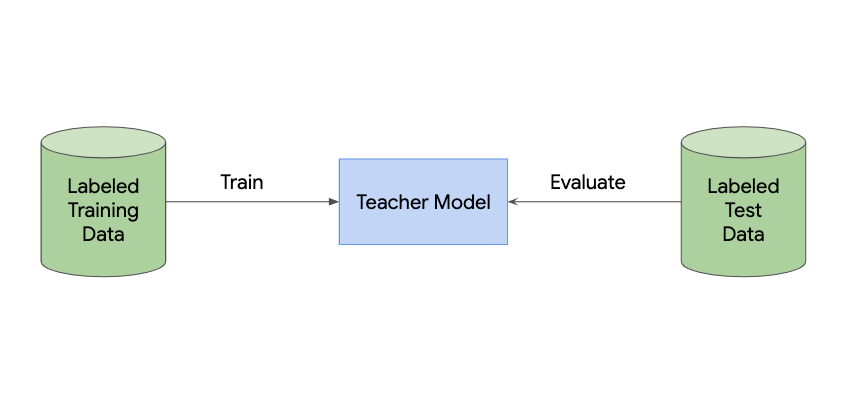}
  \caption{Teacher Model trained on labeled training data, and evaluated with labeled test data.}
  \label{teacher-model}
\end{figure}

\begin{figure}[h]
  \centering
  \includegraphics[width=0.6\linewidth, height=4.0cm]{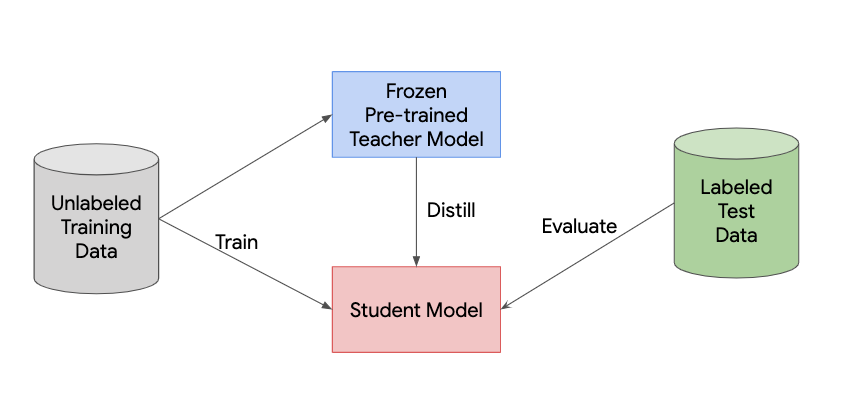}
  \caption{Student Model trained on unlabeled training data using the pre-trained teacher model, and evaluated on the labeled test data.}
  \label{student-model}
\end{figure}

\begin{figure}[h]
  \centering
  \includegraphics[width=0.6\linewidth, height=4.0cm]{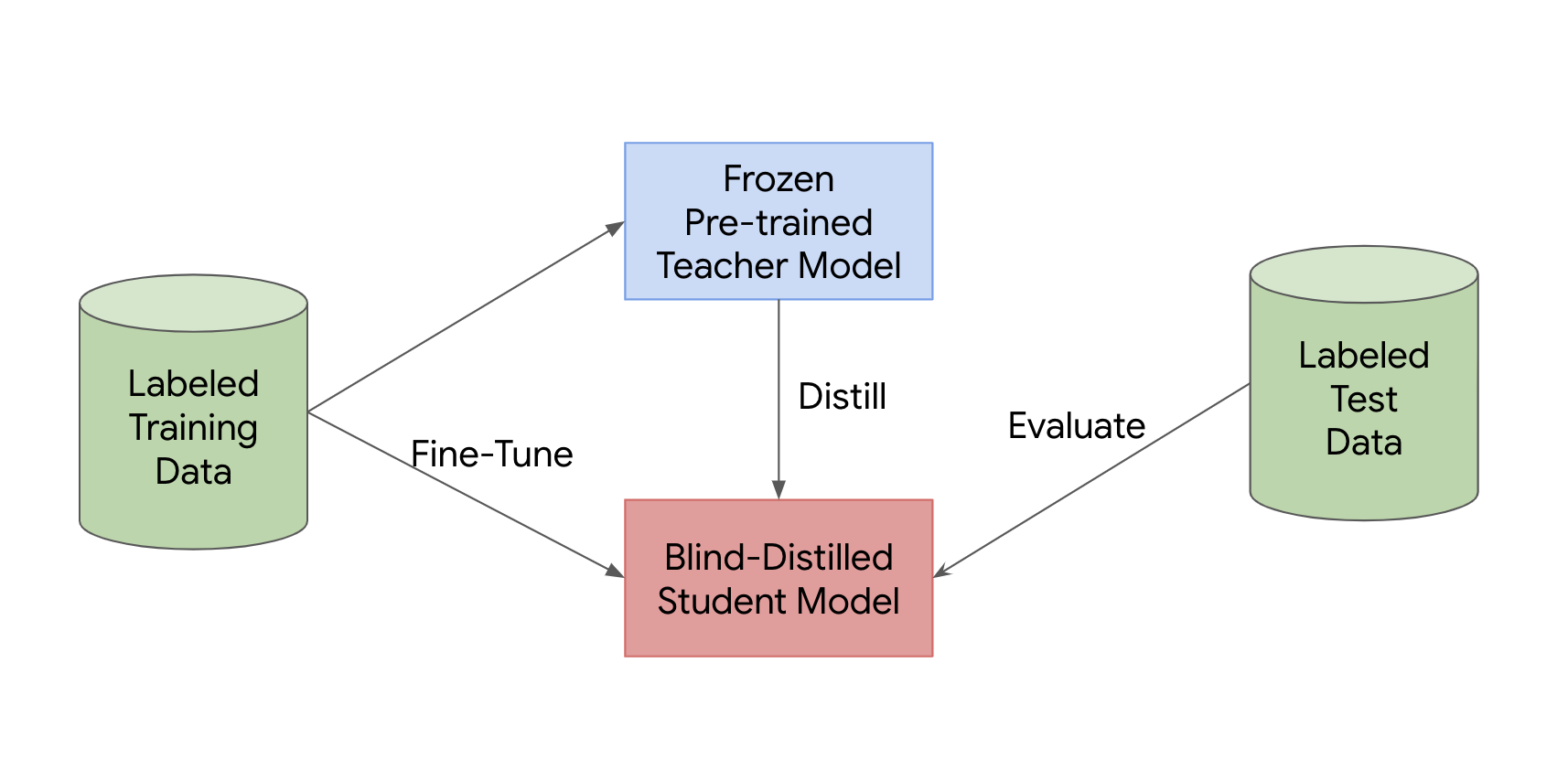}
  \caption{Fine-tuning the Blind-Distilled model with (a part of) the labeled data.}
  \label{fine-tune}
\end{figure}

To understand the intuition and ground our hypothesis in a real-word example, let us consider the task of labeling an image of a dog into one of hundred or so breeds. We can use a distillation setup with teacher and student models to learn an image classifier model. Given an unlabeled dataset of all animals excluding dogs, can a teacher model help the student distinguish that certain animals look closer to certain dog / breeds? As an example, an {\it Arctic Wolf} might look similar to a {\it Husky / Alaskan Malamute}. While, a lot of other animals might match some selective attributes of specific breeds (e.g., size, fur texture, color, etc.), others might not match any characteristics, and hence serve as hard negatives.

If the teacher model is resilient to out-of-distribution data, and can be useful for a student model, it will have several applications such as:
\begin{itemize}
    \item Using readily available unlabeled data for improving model quality, as a complementary step to the standard supervised training.
    \item Checking the robustness of a model to being replicated, when its predictions exposed through Cloud or a similar API.
\end{itemize}

In this paper, we propose a method for testing our hypothesis and demonstrate results via the following:

\begin{enumerate}
  \item Explore training student model(s) in an unsupervised setting, on an unlabeled dataset ($U$) which is completely different from the original labeled dataset ($L$) used to train the teacher model. 
  \item Demonstrate via experimental results on multiple datasets and also compare it to training directly on the original labeled dataset.
  \item Show visualizations of embeddings of examples from both the labeled and unlabeled datasets, using the teacher model which was directly trained on the labeled dataset.
\end{enumerate}

\section{Related Work}
It seems that the first approach towards transferring knowledge between networks was introduced by Caruna et al. \cite{Caruana06}, but the work was limited to shallow networks. Knowledge Distillation introduced by Hinton et al. \cite{Hinton14} allowed compressing wide and deep ensemble networks by training a smaller network that mimicked the prediction of the bigger ensemble.

Several papers point out the applications of Knowledge Distillation towards learning compact models. Chen et al. \cite{Chen17} use it to train smaller Object Detection models. Luo et al. \cite{Luo16} use it to optimize Face Recognition models. 

There is existing literature in the area of working with different data distributions for training a model. Abbasi et al. \cite{Abbasi18} explore that out-distribution samples can be used to augment a convolutional work to make it more resistant to adversarial attacks. \cite{Wang19} uses Distillation, but to construct a dataset that can be used to fine-tune a model trained on a different dataset, to predict examples based on the labeled dataset. \cite{Motiian17} provide a few-shot method for domain translation based on adversarial techniques.

Radosavovic et al. \cite{Radosavovic18} ensemble predictions over several unlabeled images during training (and optionally inference) to improve the accuracy of object detection models, but they use data from the same distribution which is the COCO labeled and unlabeled datasets. In contrast, our method does not require ensembling of transformed examples (or any other usecase-specific transforms), and we demonstrate our results on unlabeled data obtained from different sources and in some cases, with a very visually recognizable different distribution.

\section{Method}
The original training and test datasets are referred to as the \textit{labeled} $(L)$ training and test datasets. Let us denote them as ($X_{L}^{(train)}$, $Y_{L}^{(train)}$) and ($X_{L}^{(test)}$, $Y_{L}^{(test)}$) respectively, where $X$ refers to the input examples, and $Y$ refers to the corresponding labels.

We also assume that we have access to an ideally large \textit{unlabeled} $(U)$ dataset, which belongs to the same modality as our labeled dataset and drawn from a related, but not necessarily the same distribution. Even if this dataset does have labels, we would ignore them. Let us denote this dataset as $X_{U}$.

\subsection{Blind Distillation}

Next, we describe our method for training a teacher/student combination for the proposed distillation scenario (see Figure~\ref{blind-dist} for the flowchart).

\begin{enumerate}
    \item Train a teacher model $T$, using ($X_{L}^{(train)}$, $Y_{L}^{(train)}$) as training data, and ($X_{L}^{(test)}$, $Y_{L}^{(test)}$) as test data.
    \item Freeze the teacher model, and let $T(x)$ return the logits of the teacher model for the input $x$.
    \item Train a student model $S$:
        \begin{enumerate}
            \item Compute the logits of $T$ on $X_{U}$.  $Y_{T}\ =\ T(X_{U})$.
            \item Similarly, compute the logits of $S$ on $X_{U}$.  $Y_{S}\ =\ S(X_{U})$.
            \item Minimize the cross-entropy loss, $CE(Y_{S}, Y_{T})$. 
            \item Repeat (a)-(c) until convergence.
            \item Compute the accuracy of the model using ($X_{L}^{(test)}$, $Y_{L}^{(test)}$).
        \end{enumerate}
\end{enumerate}

Note that the student model $S$ trained with this method has not seen \textit{any} examples from the labeled set $X_{L}$ during training. The only source of information it has about the distribution of $X_{L}$ is through the teacher model $T$, and the information the teacher model can extract from $X_{U}$. Hence, we refer to this method as \textbf{Blind Distillation}.

\subsection{Fine-Tune with Labeled Data}

The unlabeled data being used in the Blind Distillation phase will very likely have a different distribution as compared to the labeled data. Since the student model is finally evaluated on the labeled data, we can improve its using a part of the original labeled dataset since this is available. 

\begin{enumerate}
    \item Train the student model $S$, with the Blind Distillation method above.
    \item Create a subset of the labeled training dataset with $n$ samples per class $(L')$. Let us denote this dataset as ($X_{L'}^{(train)}$, $Y_{L'}^{(train)}$).
    \item Fine tune the model $S$ by distilling from the teacher model $T$ using examples from $X_{L'}^{(train)}$. Let us refer to the model trained by this fine-tuning as $S'$.
\end{enumerate}

During our experiments, we found that fine-tuning with a low learning rate (when compared to the Blind Distillation phase) with a steady decay, helps keep the student model's loss within a reasonable range. This is intuitive because the student model has been trained to expect examples from the unlabeled dataset's distribution. Introducing examples from the labeled distribution without a careful learning rate schedule would suddenly lead to a large gradient flow. 

\section{Experiments}
We experimented with our method on standard image classification tasks. In all of our experiments, we set the distillation temperature, $t = 1.0$.

\subsection{Datasets}
We used the following datasets for our experiments:
\begin{enumerate}
    \item \textbf{MNIST}: MNIST \cite{MNIST} is a popular dataset of handwritten digits, used as a benchmark in image classification tasks.
    \item \textbf{Fashion-MNIST}: Fashion-MNIST is a dataset of fashion articles that is intended to be a drop-in replacement for the original MNIST dataset for benchmarking machine learning algorithms \cite{FashionMNIST}. Each example in the dataset is a grayscale image of size 28x28, and belongs to one of 10 classes (the same as MNIST), however the classes are completely different between the two datasets.
    \item \textbf{CIFAR-10 \& CIFAR-100}: The CIFAR-10 dataset consists of 60000 32x32 colour images belonging to 10 classes. CIFAR-100 is similar to CIFAR-10, except that each image belongs to one of 100 classes. CIFAR-100 can be considered similar to CIFAR-10 in the taxonomy that they represent \cite{CIFARDatasets}. However, none of the classes in CIFAR-10 overlap with any class in CIFAR-100.
    \item \textbf{Caltech-256}: The CalTech-256 dataset is a collection of ${\sim}$ 30,000 images spread across 256 classes \cite{Caltech256Dataset}. Each image belongs to one of the 256 classes.
    \item \textbf{OpenImages}: Open Images \cite{OpenImagesDataset} has ${\sim}$ 9M images spanning ${\sim}$ 20,000 classes. The images often contain complex scenes with several objects.
    \item \textbf{ImageNet}: ImageNet \cite{ImageNetDataset} is a popular image classification dataset having ${\sim}$ 1M images spanning ${\sim}$ 1000 classes. Each image contains one image.
\end{enumerate}

\subsection{Infrastructure}
We used Google Colab \cite{Colab} for experimenting with MNIST, Fashion MNIST, and CIFAR datasets, using one GPU at a time. For experimenting with Caltech-256 and OpenImages we used an internal Machine Learning training platform using 4 GPUs at a time.

In the following subsection, we have broken down the experimentation section based on the labeled and unlabeled dataset pairs.

\begin{figure}[h]
  \centering
  \includegraphics[width=0.75\linewidth, height=4.20cm]{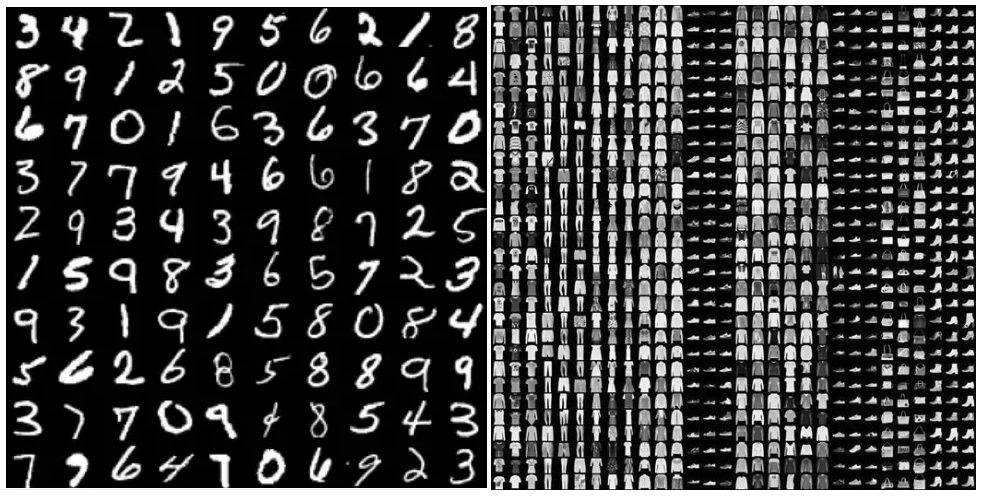}
  \caption{Comparison between the MNIST and the Fashion MNIST dataset. Source: \cite{MNISTComparisonImg}}
  \label{mnist-fmnist-dataset}
\end{figure}

\subsection{MNIST Classification Results: <$L=\ $MNIST, $\ U=\ $Fashion MNIST>}

We use MNIST as the labeled dataset, and Fashion MNIST as the unlabeled dataset. Figure~\ref{mnist-fmnist-dataset} shows the visual difference between the two datasets. Table~\ref{mnist-fmnist} shows results for this experimental setting.

We experiment with two architectures:
\begin{enumerate}
    \item CNN: Two convolutional layers, with batch-normalization and max-pooling layers between them, and one fully connected layer. Total: $\sim$ 1.12M params.
    \item MLP: Three fully connected layers. Total: $\sim$ 85,000 params.
\end{enumerate}

Table~\ref{mnist-fmnist} shows the accuracy of the teacher model (column 1), as compared to the accuracy of the student model when blind-distilled using Fashion MNIST and the teacher model (column 2). The third column shows the accuracy when the student model is fine-tuned using the MNIST labeled data.

When using CNN-based teacher and student models, the student gets 96.2\% accuracy when compared to the teacher's 99.5\% accuracy, without using any MNIST examples.

Similarly using MLP-based teacher and student models, the student gets 84\% accuracy when compared to the teacher's 97.69\% accuracy, again without using any MNIST examples.

We will further discuss the performance differences between the combination of the two model architectures in the Discussion section.


Another interesting observation is that when the teacher and student model capacities are lower, it leads to lower student model accuracy. This implies that the student is learning a non-trivial function, and both the teacher model accuracy as well as the student model’s own capacity to learn are important.

\begin{table}
  \caption{Distillation Results on MNIST \& Fashion MNIST}
  \label{mnist-fmnist}
  \centering
  \begin{tabular}{lll}
    \toprule
    Teacher Model         & Blind Distillation  & Blind Distillation \\
    Accuracy ($T$)          & Accuracy ($S$)        & \& Fine-Tune ($S'$) \\
    \midrule
    99.5\% (CNN)            & \textbf{96.2\%} (CNN)    & 99.6\%     \\
    97.69\% (MLP)           & 84\% (MLP)               & 97.95\%     \\
    99.5\% (CNN)            & 68.60\% (MLP)            & 98.16\%     \\
    97.69\% (MLP)           & 89.45\% (CNN)            & 99.26\%     \\
    \bottomrule
  \end{tabular}
\end{table}

\subsection{CIFAR Image Classification Results: <$L=\ $CIFAR-10, $\ U=\ $CIFAR-100>}
For these experiments, we use the same model architecture for both the teacher and the student.

The architecture for both the teacher and the student is a simple Convolutional Neural Net with 6 convolutional layers, and one feed forward layer. We add batch-normalization layers and dropout layers to help improve accuracy.

The student model reaches an accuracy of 79.5\%, without having seen any examples from the CIFAR-10 dataset. Fine-tuning with the labeled dataset improved the accuracy to 86.37\%, which was better than the teacher model's 84.73\%.

\begin{table}
  \caption{Distillation Results on CIFAR-10 \& CIFAR-100}
  \label{cifar10-cifar100}
  \centering
  \begin{tabular}{lll}
    \toprule
    Teacher Model         & Blind Distillation  & Blind Distillation \\
    Accuracy ($T$)          & Accuracy ($S$)        & \& Fine-Tune ($S'$) \\
    \midrule
    84.73\% (CNN)         & 79.5\% (CNN)        & 86.37\%     \\
    \bottomrule
  \end{tabular}
\end{table}

\subsection{Image Classification Results with More Classes: <$L{=}$CalTech-256, $\ U=\ $OpenImages>}
In this experiment, both the teacher and student models were based on the MobileNet v2 architecture \cite{MobileNetV2}. 

This is a more challenging experiment, as Caltech-256 is an ImageNet like dataset with a large number of classes. OpenImages and covers a wide range of concepts, with ~8 different classes present on an average in the same image \cite{OpenImagesDataset}. We believe that a well trained teacher model would be able to extract  

Note that fine-tuning with just 10 and 20 samples / class on the CalTech256 dataset improved the accuracy of the student model from 64.0\% to 71.6\% and 74.46\% respectively, doing significantly better than the original teacher model. Using the entire dataset to fine-tune the student model leads to an even bigger jump to 76.49\% accuracy.

\begin{table}[h]
  \caption{Distillation Results on Caltech-256 \& OpenImages}
  \label{caltech256-openimages}
  \centering
  \begin{tabular}{lllll}
    \toprule
    Teacher Model         & Blind Distillation  & \multicolumn{3}{c}{Blind Distillation  \&} \\
    Accuracy ($T$) & Accuracy ($S$)  & \multicolumn{3}{c}{Fine-Tune Accuracy ($S'$)} \\
    \cmidrule(r){3-5}
    &   & $n=10$ & $n=20$ & $n=$ALL \\
    \midrule
    68.20\% & 64.0\% & 71.60\% & 74.46\% & 76.49\%    \\
    \bottomrule
  \end{tabular}
\end{table}

\subsection{Image Classification Results with Even More Classes: <$L{=}$ImageNet, $\ U=\ $OpenImages>}
In this experiment as well, both the teacher and student models were based on the MobileNet v2 architecture \cite{MobileNetV2}.

\begin{table}[h]
  \caption{Distillation Results on ImageNet \& OpenImages}
  \label{caltech256-openimages}
  \centering
  \begin{tabular}{lllll}
    \toprule
    Teacher Model         & Blind Distillation  & \multicolumn{3}{c}{Blind Distillation  \&} \\
    Accuracy ($T$) & Accuracy ($S$)  & \multicolumn{3}{c}{Fine-Tune Accuracy ($S'$)} \\
    \cmidrule(r){3-5}
    &   & $n=20$ & $n=60$ & $n=$ALL \\
    \midrule
    72.0\% & 55.6\% & X\% & Y\% & 67.75\%    \\
    72.0\% & 59.5\% & X\% & Y\% & 68.08\%    \\
    \bottomrule
  \end{tabular}
\end{table}

\section{Discussion}


In this section we further analyse the results of our experiments, w.r.t. the embeddings of the labeled and unlabeled data, effect of teacher and student model architecture, etc.

\subsection{Joint embedding of examples from $L$ and $U$}

\begin{figure}[h]
  \centering
  \includegraphics[scale=0.5]{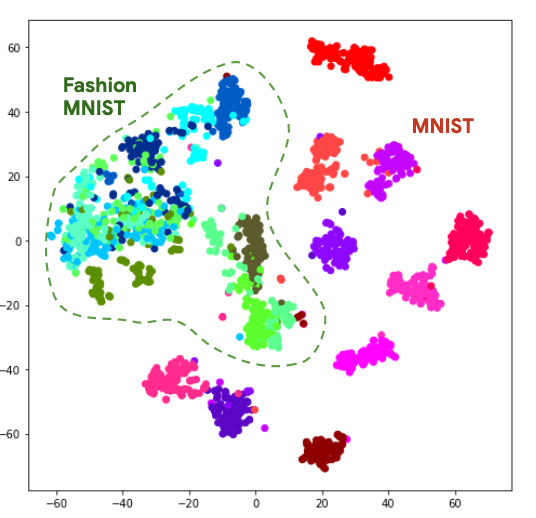}
  \caption{t-SNE plot of the embeddings of MNIST and Fashion MNIST images, as extracted from a teacher model trained on MNIST data.}
  \label{tsne}
\end{figure}

Figure~\ref{tsne} is the t-SNE plot of embeddings of MNIST and Fashion MNIST images (1000 images from either dataset) produced by the teacher model. The shades of green and blue are images belonging to Fashion MNIST. Shades of red, pink and purple belong to MNIST. Varying the t-SNE learning rate in the [100.0, 300.0] range and perplexity in the [20.0, 50.0] range gave similar plots.

We can see distinct clusters of MNIST classes, as the teacher model learnt the manifold that the MNIST images occupy.

We can also observe that the Fashion MNIST images are not spread randomly, although certain MNIST images are clearly on the Fashion MNIST side of the boundary. Certain clusters for example the bright and dark green dots are closer to some MNIST image clusters. Similarly the navy blue cluster at the top is closer to the dark red MNIST cluster. Examples drawn from these Fashion MNIST classes would therefore make the teacher predict the examples to be more likely to belong to the MNIST classes they are close to, than to the classes they are far apart from.

\subsection{Student Model Architecture}
In the MNIST \& Fashion MNIST experiment, we experiment with two different model architectures. A simple convolutional neural net (CNN) architecture outperformed a multi-layer perceptron (MLP) network. This was intuitive, and expected. All other things held constant, we find that if the student network had more capacity to learn, it performed better in our setup.

\subsection{Teacher Model Architecture}
We see an interesting phenomenon in the Blind Distillation phase when changing the teacher model architecture while keeping the student model architecture same. If the teacher was an MLP, and the student was a CNN, it was \textit{less} accurate than was when the teacher was a CNN as well. Our hypothesis is that a CNN teacher model architecture extracted features that were different from what an MLP extracted, hence leading to a dip in the performance. Similarly, when the teacher was a CNN, but the student was an MLP, it performed quite worse than when the teacher was an MLP too.

However, when these blind-distilled models were fine-tuned on the actual datasets, we found that both the performance dips were resolved. When the MLP model was being fine-tuned by a CNN model, its performance improved to 98.16\%, which is better than either vanilla supervised training or blind-distillation followed by fine-tuning with an MLP.

Similarly, the performance of the CNN model being fine-tuned by an MLP improved its accuracy from 89.45\% to 99.26\%. This is lower than the CNN model being trained by itself / distilled from another CNN model, but this is expected since the teacher model has $\sim$ 10x less parameters.

\subsection{Accuracy of the Teacher}

It is intuitive to expect a more accurate teacher to lead to a better trained student, and we saw that pattern across the different datasets. In addition, we saw that while keeping the teacher model architectures the same, the cross entropy loss at convergence was higher for the student distilled from the more accurate teacher model. This hints at the case that more accurate teachers will teach more complex functions. 

\subsection{Comparison with Transfer Learning}
Transfer Learning can transfer knowledge between labeled domains. Our work demonstrates the efficacy of distillation when used in an unlabeled domain. 

As such, our methods are complementary to Transfer Learning, since the Teacher and/or the student can be initialized with a transfer-learnt checkpoint. 

\section{Conclusion}
In this paper, we saw that a teacher model used in a distillation setup, captured relationships between classes, even outside the original dataset. For example, while the MNIST and Fashion MNIST datasets look very different, using Blind Distillation, we were able to train a model to 96\% accuracy, which hasn't seen \textit{any} MNIST images. Using the t-SNE plots between MNIST and Fashion MNIST images, we visually demonstrated why distillation is surprisingly effective in such a constrained setting.

Similarly, the Caltech-256 and the OpenImages dataset differ a lot in terms of the number of classes and distribution of images. Blind Distillation by itself brings us to 64.0\% accuracy as compared to the 68.20\% accuracy of the teacher. Fine-tuning using the labeled examples leads to a $\sim$ 8.30\% absolute accuracy improvement. This is promising because it is relatively easier and cheaper to collect large amounts of unlabeled data, than it is to get the same amount of labeled data. Leveraging the readily available unlabeled data in conjunction with existing supervised learning on the labeled data, offers a scenario which is best of both the worlds.

From the perspective of model security research, we demonstrate that given access to a model's logits, it is easy to replicate the given model's performance without access to the original data. This is a problem worth investigating in its own right.

With this paper, we hope we can further the understanding of Knowledge Distillation, and inspire its applications which leverage unlabeled data to improve model quality. 

\small
\bibliographystyle{unsrt}

\begin{thebibliography}{9}
\bibitem{Hinton14}
G. Hinton, O. Vinyals, J. Dean. Distilling the Knowledge in a Neural Network, NeurIPS Deep Learning Workshop 2014.

\bibitem{Caruana06}
C. Bucilua, R. Caruana, A. Niculescu-Mizil. Model compression. KDD 2006.

\bibitem{Chen17}
G. Chen, W. Choi, X. Yu, T. Han, M. Chandraker. Learning Efficient Object Detection Models with
Knowledge Distillation. NeurIPS 2017.

\bibitem{Cheng19}
Y. Cheng, D. Wang, P. Zhou. A Survey of Model Compression and Acceleration for Deep Neural Networks. IEEE Signal Processing Magazine Special Issue on Deep Learning for Image Understanding, 2019.

\bibitem{Luo16}
P. Luo, Z. Zhu, Z. Liu, X. Wang, X. Tang. Face Model Compression by Distilling Knowledge from Neurons. AAAI 2016.

\bibitem{Abbasi18}
M. Abbasi, C. Gagne. Out-distribution Training Confers Robustness To Deep Neural Networks. ICLR Workshop Track 2018.

\bibitem{Wang19}
T. Wang, J. Zhu, A. Torralba, A. A. Efros. Dataset Distillation. arXiv:1811.10959, 2019.

\bibitem{Motiian17}
S. Motiian, Q. Jones, S. M. Iranmanesh, G. Doretto. Few-Shot Adversarial Domain Adaptation. NeurIPS 2017.

\bibitem{Radosavovic18}
I. Radosavovic, P. Dollar, R. Girshick, G. Gkioxari, K. He. Data Distillation: Towards Omni-Supervised Learning. CVPR 2018.

\bibitem{MNIST}
Y. LeCun, C. Cortes, C. J. C. Burges. \textit{The MNIST Dataset Of Handwritten Digits}.

\bibitem{FashionMNIST}
H. Xiao, K. Rasul, R. Vollgraf. Fashion-MNIST: a Novel Image Dataset for Benchmarking Machine Learning Algorithms. arXiv:1708.07747, 2017.


\bibitem{CIFARDatasets}
A. Krizhevsky. \url{https://www.cs.toronto.edu/~kriz/cifar.html}.

\bibitem{Caltech256Dataset}
G. Griffin, A. D. Holub, P. Perona. The Caltech 256. Caltech Technical Report.

\bibitem{ImageNet}
J. Deng, W. Dong, R. Socher, L.-J. Li, K. Li, L. Fei-Fei,. ImageNet: a large-scale hierarchical image database. CVPR, 2009.

\bibitem{Colab}
Google Colaboratory. \url{https://colab.research.google.com}.

\bibitem{OpenImagesDataset}
A. Kuznetsova, H. Rom, N. Alldrin, J. Uijlings, I. Krasin, J. Pont-Tuset, S. Kamali, S. Popov, M. Malloci, T. Duerig, and V. Ferrari. The Open Images Dataset V4: Unified image classification, object detection, and visual relationship detection at scale. arXiv:1811.00982, 2018.

\bibitem{ImageNetDataset}
J. Deng, W. Dong, R. Socher, L.-J. Li, K. Li, and
Fei-Fei, L. ImageNet: a large-scale hierarchical image database. CVPR, 2009.

\bibitem{MobileNetV2}
M. Sandler, A. Howard, M. Zhu, A. Zhmoginov, L. Chen. MobileNetV2: Inverted Residuals and Linear Bottlenecks. CVPR 2018.

\bibitem{MNISTComparisonImg}
The Fashionable “Hello World” of Deep Learning. \url{https://medium.com/analytics-vidhya/the-fashionable-hello-world-of-deep-learning-8b9e3d60a37c}
\end{thebibliography}

\end{document}